\documentclass[a4paper,fleqn]{cas-dc}

\usepackage[numbers]{natbib}
\usepackage{longtable}
\usepackage{tabularx}
\usepackage{algorithm}
\usepackage{algorithmic}
\usepackage{float}
\usepackage{arydshln}
\usepackage[protrusion=true,expansion=false]{microtype}
\usepackage{placeins}
\usepackage{balance}

\renewcommand{\arraystretch}{1.20}
\newcolumntype{Y}{>{\raggedright\arraybackslash}X}
\ExplSyntaxOn
\RenewDocumentCommand \printorcid { }
  {
    \seq_if_empty:NF \g_stm_orcid_seq
      {
        \group_begin:
        \tex_let:D \thefootnote \relax \footnotetext
          {
            \raggedright
            \textsc{orcid}(s):\c_space_token
            \seq_use:Nn \g_stm_orcid_seq { ;~ }
          }
        \group_end:
      }
  }
\ExplSyntaxOff

\begin{document}
\let\WriteBookmarks\relax
\def\floatpagepagefraction{1}
\def\textpagefraction{.001}
\hypersetup{hypertexnames=false}

\shorttitle{}
\shortauthors{Du et al.}

\title[mode=title]{Task-Driven 3D Printability Assistance via Geometry- and Knowledge-Grounded LLM Reasoning}

\author[1]{Zhaoda Du}
\ead{zhaoda_du@mines.edu}

\author[1]{Qiaojie Zheng}
\ead{zheng@mines.edu}

\author[1]{Xiaoli Zhang}
\cormark[1]
\ead{xlzhang@mines.edu}

\affiliation[1]{organization={Colorado School of Mines},
            addressline={1500 Illinois St.},
            city={Golden},
            postcode={80401},
            state={CO},
            country={USA}}

\cortext[1]{Corresponding author.}

\begin{abstract}
Printability assessment in additive manufacturing is typically conducted at the geometry level before printing to determine whether a computer-aided design (CAD) model or stereolithography (STL) file can be successfully fabricated. Task suitability, in contrast, is usually evaluated after printing to determine whether the fabricated part satisfies the requirements of its intended use. As a result, for non-expert users to print functional parts, unsuitable material or process choices may only be identified after fabrication, leading to repeated printing, material waste, and user frustration. To address this challenge, this paper leverages the reasoning and language-understanding capabilities of large language models (LLMs), while grounding the reasoning with geometry evidence and structured material/printer knowledge to generate reliable pre-print recommendations. Given a stereolithography (STL) model and a natural-language task description, the framework generates a structured recommendation covering printability, material choice, process parameters, design guidance, risks, and explanations. We evaluate the framework on focused STL benchmark scenarios with novice-style task descriptions. The proposed method achieves 75.0\% printability over 96 physical validation trials, with 88.9\% task suitability among successfully printed samples. It also improves Gemini 2.5 Flash-Lite material-selection accuracy from 37.5\% under pure LLM to 90.0\%. Expert evaluation further shows improved report quality, while post-print feedback improves recommendations on selected problematic cases. These results suggest that user task intent, geometry evidence, and structured material knowledge are all important for reliable task-driven printability assistance.
\end{abstract}

\begin{keywords}
Additive manufacturing \sep 3D printability   \sep Large language models \sep Knowledge graph 
\end{keywords}

\maketitle

\section{Introduction}
\begin{figure*}[width=\textwidth,pos=t]
    \centering
    \includegraphics[width=\linewidth]{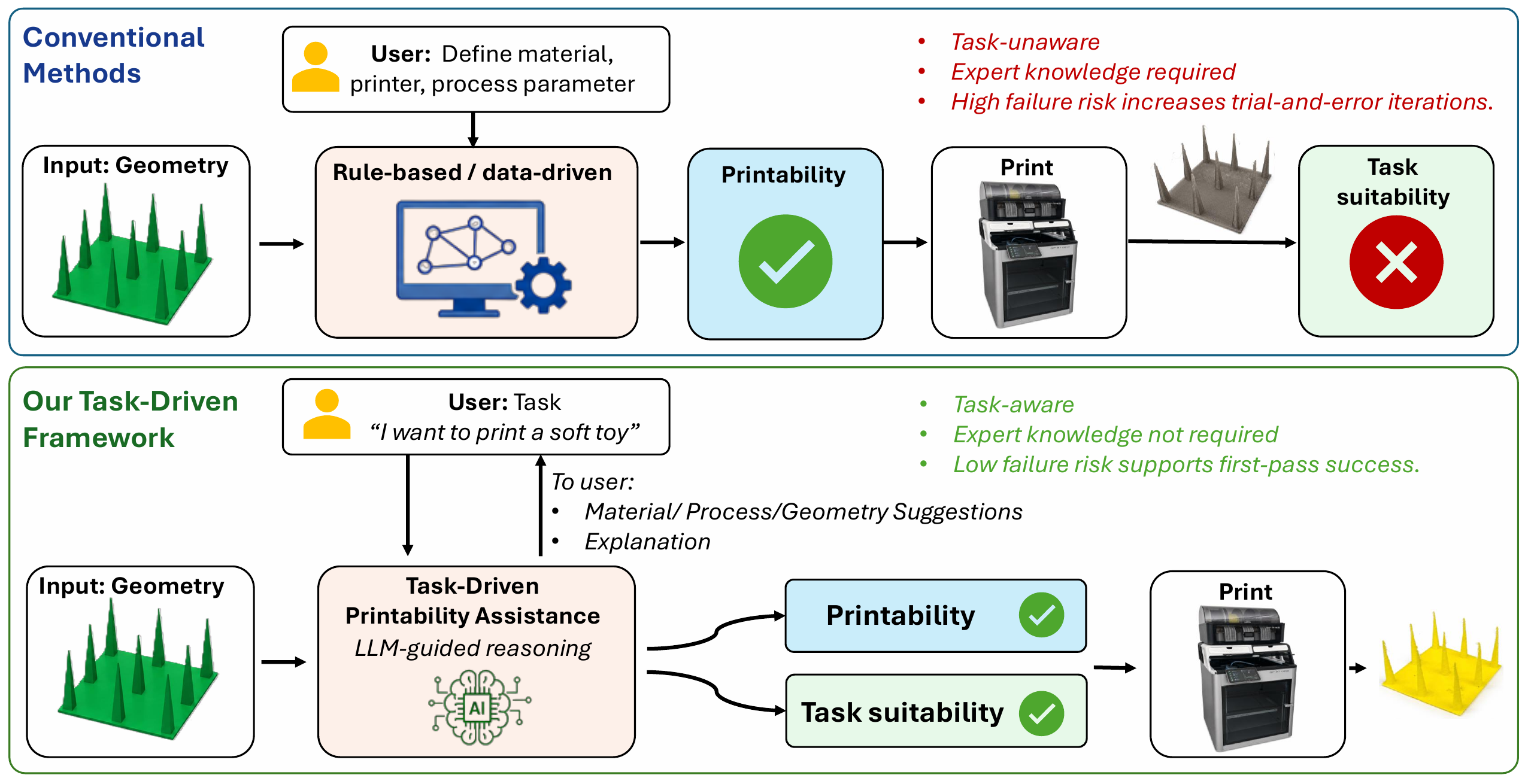}
    \caption{
    Comparison between conventional printability evaluation and the proposed task-driven printability assistance framework. }
    \label{fig:Overall}
\end{figure*}

Assessing printability before fabrication is important in Additive Manufacturing (AM) for time and cost reduction. In current 3D printing workflows, printability is typically understood at the geometry level as whether an input computer-aided design (CAD) or stereolithography (STL) model can be fabricated under a given printer, material, and process setup without major appearance failures. However, as AM has become increasingly mature, it is no longer limited to hobbyist prototyping but is now widely used to fabricate replacement parts, custom fixtures, and other application-dependent objects. In these applications, geometric printability alone is insufficient, because a printed part must also satisfy the requirements of its intended use.

Simultaneously considering geometric printability and task suitability is challenging because the two objectives operate at different levels and are not independent \cite{goala2024selection,altiparmak2024suitability}. Geometric printability concerns whether the input geometry can be fabricated reliably under a given setup \cite{fudos2020characterization}, whereas task suitability concerns whether the printed part can satisfy the requirements of its intended use after fabrication. At the same time, these objectives are often coupled through material and process choices \cite{booth2017design}: a material or setting that better supports the intended use may introduce new fabrication difficulties, while a setup that improves geometric printability may reduce suitability for the target application \cite{uz2018integrated,altiparmak2024suitability,goala2024selection}.

This challenge is not well addressed by existing automated printability assessment methods \cite{fudos2020characterization,lu2016towards,mycroft2020data,trovato2024decision}. Although these methods can identify geometry-related risks such as overhangs,  wall-thickness issues, or mesh defects, they do not infer the user’s intended use or jointly recommend material and process choices under task requirements. For example, a geometry-level checker may judge a model as printable with PLA, but the printed part may soften or deform if the intended task requires use under heat exposure.

Task suitability has often been addressed through expert empirical initial material/process selection combined with post-print evaluation and iteration \cite{pradel2018investigation,frank1995expert}. Consequently, task suitability is highly dependent on expert knowledge and often requires repeated printing before a suitable part is obtained.

Although prior work has recognized printability and task suitability as related but distinct objectives~\cite{altiparmak2024suitability}, existing studies do not jointly reason about them for a specific input geometry and user task before printing. In practice, these two aspects are often handled sequentially rather than jointly, as illustrated in Figure~\ref{fig:Overall}. This workflow is not fully automated and usually requires expert intervention to select materials, adjust process parameters, interpret printability risks, and revise the design after failures. As a result, it is inefficient, failure-prone, and highly dependent on expert knowledge~\cite{pradel2018investigation,frank1995expert}.

\begin{figure*}[width=\textwidth,pos=!t]
    \centering
    \includegraphics[width=\linewidth]{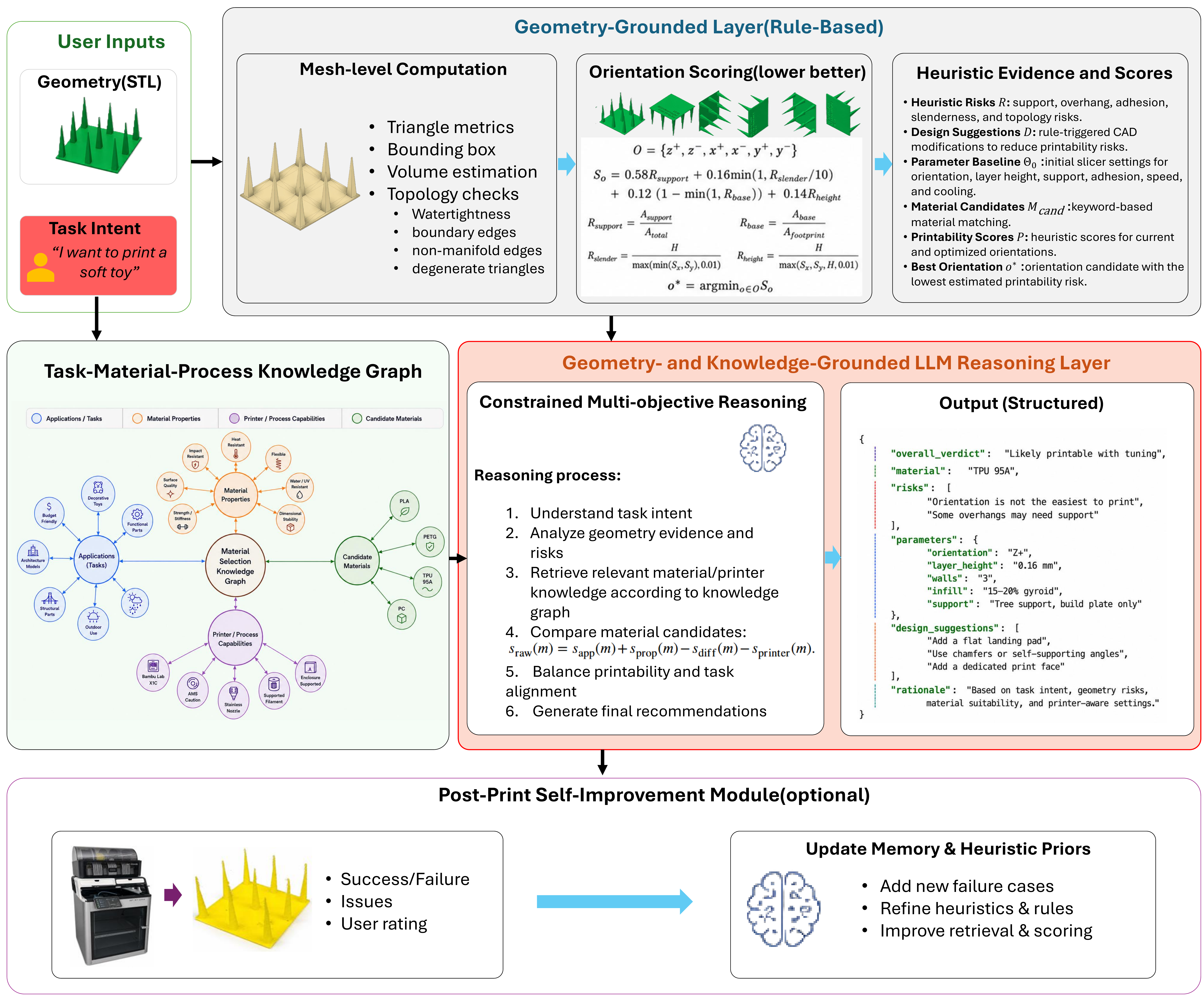}
    \caption{Overview of the proposed task-driven printability assistance framework.}
    \label{fig:Figure2}
\end{figure*}

To address this joint reasoning gap, we leverage the reasoning and language-understanding capabilities of LLMs to connect ambiguous user task intent with geometry, material, and process requirements. Specifically, we develop a geometry- and knowledge-grounded LLM framework for task-driven printability assistance, as shown in Figure~\ref{fig:Figure2}. Given a natural-language task intent and an input STL model, the framework first converts the geometry into a structured evidence package through rule-based mesh-level computation, orientation scoring, heuristic risk generation, parameter baselines, and material candidates. It then combines this geometry-grounded evidence with a structured Task-Material-Process knowledge graph and uses constrained multi-objective LLM reasoning to generate a structured recommendation report. The report includes an overall printability assessment, task-relevant risks, material and process-parameter suggestions, design guidance, and explanations. An optional post-print self-improvement module records observed outcomes and user ratings to update memory, heuristics, and future scoring.

The main contributions of this work are summarized as follows:
\begin{sloppypar}
\begin{enumerate}
    \item \textbf{A task-driven formulation of printability assistance.}
    We formulate printability assistance as a task-dependent problem, where the same geometry may require different material, parameter, design, and risk recommendations depending on the user's intended task.

    \item \textbf{A geometry- and knowledge-grounded LLM framework for task-driven printability assistance.}
   We propose a framework that combines geometry evidence, a structured Task-Material-Process knowledge graph, and constrained LLM reasoning to generate task-driven printability reports with risks, explanations, and material/parameter/design suggestions.

    \item \textbf{Evaluation methodology for task-driven printability assistance.}
We evaluate the framework across multiple LLM backbones using novice-style task descriptions and same-geometry different-task cases, with evaluation organized around geometry-level printability evaluation, task suitability evaluation, material selection accuracy, and expert evaluation of report quality. We further assess the self-improvement module on selected problematic cases.
\end{enumerate}
\end{sloppypar}

\section{Related Work}
\label{sec:related_work}

\subsection{Rule-Based Printability Assessment}
\label{subsec:rule_based_printability}

Traditional printability assessment methods often evaluate whether a CAD or STL model satisfies geometry- or process-specific constraints under a predefined manufacturing setup. Automated printability checkers typically extract geometric features and compare them with printer constraints such as build volume, minimum feature size, wall thickness, overhang limits, support requirements, and mesh validity. For example, Lu formulated printability checking as a rule-based verification problem over object features and printer profiles~\cite{lu2016towards}, while Fudos et al. characterized 3D printability through geometry- and technology-dependent printability scores~\cite{fudos2020characterization}. Related work on support generation, model decomposition, build-orientation planning, and multi-axis support-free fabrication further shows that printability depends on orientation, support strategy, and process planning rather than geometry alone~\cite{jiang2019optimization,vanek2014clever,luo2012chopper,han2025support,li2024supportless,guo2024design}.

These methods provide interpretable and reproducible manufacturability evidence, which is important for practical decision support. However, they are generally setup-first: the printer, material, orientation objective, and process constraints are assumed before assessment. As a result, they can identify geometry-related risks, but they usually do not infer the user's task intent or adapt material, parameter, and design recommendations to task suitability requirements.

\subsection{Data-Driven Printability Prediction}
\label{subsec:data_driven_printability}
Data-driven methods extend rule-based assessment by learning relationships between geometry, process conditions, and print outcomes. Geometry-centered approaches use supervised machine-learning models, decision-tree methods, or neural networks to learn from CAD features, local descriptors, STL-derived geometric representations, or printed test artifacts~\cite{mycroft2020data,trovato2024decision,henn2025evaluating}. For example, Trovato and Cicconi used a decision-tree approach for early evaluation of 3D models~\cite{trovato2024decision}, while Henn et al. trained a convolutional neural network to detect geometry-related printability issues from STL-derived inputs before printing~\cite{henn2025evaluating}.

Data-driven process-aware models further incorporate material behavior and process parameters. Hammoud et al. combined experimental characterization with machine-learning prediction for 3D concrete printing, including geometric-quality prediction, inverse parameter selection, and printability maps~\cite{hammoud2025data}. Other studies model geometric deviation under different printhead and process conditions or use process-window prediction and in-situ defect monitoring to account for process context and observed outcomes~\cite{kvrivohlavy2025influence,chen2020predicting,gobert2018application,sahar2023anomaly,fu2021situ}.

Although data-driven and process-aware models can provide useful predictions and, in some cases, parameter-selection guidance, they remain limited for task-driven assistance. They often require substantial training data, provide limited explanation for decisions, and are tied to specific machines, materials, process windows, geometry families, or quality metrics.

\subsection{LLM-Based Manufacturing Assistance}
\label{subsec:llm_manufacturing_assistance}

Large language models(LLMs) offer a useful interface for manufacturing assistance because they can interpret natural-language instructions, reason over ambiguous goals, and generate explanations understandable to non-expert users \cite{ouerghemmi2025integrating,li2026large,du2026decision}. Recent studies have explored LLMs and vision-language models for additive-manufacturing knowledge querying, defect prediction, process monitoring, and manufacturing decision support \cite{zheng2025context,zheng2026qa,jadhav2025llm,ouerghemmi2025integrating,pak2025additivellm, eslaminia2025fdm, chandrasekhar2024amgpt}. AMGPT investigates LLM-based contextual querying in additive manufacturing ~\cite{chandrasekhar2024amgpt}, while FDM-Bench evaluates LLM competence on domain-specific additive-manufacturing tasks ~\cite{eslaminia2025fdm}. AdditiveLLM studies LLM-based defect prediction in metal additive manufacturing ~\cite{pak2025additivellm}, and LLM-3D Print demonstrates that language-model agents can monitor print images, diagnose defects, and suggest or execute corrective actions during material-extrusion printing ~\cite{jadhav2025llm}. Broader reviews also discuss the integration of LLMs into digital manufacturing workflows ~\cite{ouerghemmi2025integrating}.

These studies show that LLMs can provide flexible interaction, technical-context interpretation, and human-readable explanations. However, existing LLM-based manufacturing systems focus on knowledge querying, monitoring, or defect diagnosis rather than pre-print task-conditioned assistance from an uploaded STL model. 
\section{Methodology}

The proposed framework consists of three functional layers and an optional post-print self-improvement module, as shown in Figure~\ref{fig:Figure2}. The Geometry-Grounded Layer models the task-geometry relation by combining geometry/topology analysis with task cues and expert-defined fabrication heuristics. The Task-Material-Process knowledge graph provides structured task-material-process relations by linking task requirements, material properties, and printer/process constraints. The Geometry- and Knowledge-Grounded LLM Reasoning Layer then integrates these task-geometry and task-material-process relations to perform task-driven reasoning over geometry, material, and process decisions. In this way, the framework generates recommendations that are not only geometrically printable but also aligned with the user's intended task. The optional post-print self-improvement module supports post-print refinement using observed outcomes and user feedback.
\subsection{Geometry-Grounded Layer}
\label{subsec:geometry_grounded_layer}
The Geometry-Grounded Layer converts the uploaded STL geometry \(G\) and lightweight task cues from \(T\) into a structured geometry-grounded evidence package \(E_{\mathrm{G}}\) for downstream LLM reasoning. This layer consists of three steps: mesh-level computation, orientation scoring, and heuristic evidence generation and scoring. Its output is defined as:
\begin{equation}
\begin{aligned}
\label{eq:geometry_evidence_package}
E_G
&=\mathcal{F}_{\mathrm{geo}}(G,T) \\
&=\{F_{\mathrm{mesh}},Q_{\mathrm{topo}},S_o,R,D,\Theta_0,
M_{\mathrm{cand}}^{0},o^*\}.
\end{aligned}
\end{equation}
Here, \(F_{\mathrm{mesh}}\) and \(Q_{\mathrm{topo}}\) denote mesh and topology evidence, \(S_o\) denotes orientation-dependent printability scores, \(R\) denotes heuristic risks, \(D\) denotes design hints, \(\Theta_0\) denotes an initial parameter baseline, \(M_{\mathrm{cand}}^{0}\) denotes keyword-based initial material candidates,  and \(o^*\) denotes the selected best orientation. The following subsections describe how these components are generated.

\subsubsection{Mesh-level Computation}
The mesh-level computation step generates the mesh and topology evidence 
\((F_{\mathrm{mesh}}, Q_{\mathrm{topo}})\). The uploaded STL is parsed as a triangle mesh, from which the layer computes geometric quantities such as model dimensions, surface area, and estimated volume. It also checks topology-related indicators such as watertightness, boundary
edges, non-manifold edges and degenerate triangles. These outputs provide low-level evidence about whether the uploaded geometry is valid and suitable for slicing~\cite{lu2016towards,fudos2020characterization}.
\subsubsection{Orientation Scoring}

The orientation scoring step generates the orientation-dependent score \(S_o\) and the selected orientation candidate \(o^*\). To account for orientation-dependent printability, the layer evaluates six canonical build orientations:
\begin{equation}
\mathcal{O}=\{z^+,z^-,x^+,x^-,y^+,y^-\}.
\end{equation}

For each orientation, the layer computes support-risk, base-contact, slenderness, and height-dominance indicators. These indicators are combined into a heuristic orientation score:
\begin{equation}
\begin{aligned}
\label{equation2}
S_o =
&\,0.58R_{\mathrm{support}}
+0.16\min\left(1,\frac{R_{\mathrm{slender}}}{10}\right) \\
&+0.12\left(1-\min(1,R_{\mathrm{base}})\right)
+0.14R_{\mathrm{height}} .
\end{aligned}
\end{equation}

The weighting coefficients in Eq.~\eqref{equation2} were empirically determined through preliminary experiments on a validation set of primitive geometries to balance the relative printability impacts of support burden, slenderness-induced instability, bed adhesion risk, and height dominance~\cite{frank1995expert,taufik2013role}. The four orientation-dependent indicators are defined as
\begin{equation}
\begin{aligned}
R_{\mathrm{support}} &= \frac{A_{\mathrm{support}}}{A_{\mathrm{total}}}, \quad
R_{\mathrm{base}} = \frac{A_{\mathrm{base}}}{A_{\mathrm{footprint}}},\\
R_{\mathrm{slender}} &=
\frac{H}{\max(\min(S_x,S_y),0.01)},\\
R_{\mathrm{height}} &=
\frac{H}{\max(S_x,S_y,H,0.01)} .
\end{aligned}
\end{equation}

Here, \(A_{\mathrm{support}}\) is the area of downward-facing near-horizontal surfaces, \(A_{\mathrm{total}}\) is the total surface area, \(A_{\mathrm{base}}\) is the estimated build-plate contact area, \(A_{\mathrm{footprint}}\) is the footprint bounding-box area, \(H\) is the build height, and \(S_x,S_y\) are the horizontal spans in the build-plate plane. The orientation with the lowest score is selected as the best orientation candidate:
\begin{equation}
o^*=\arg\min_{o\in\mathcal{O}}S_o .
\end{equation}

The resulting \(S_o\) and \(o^*\) are included in the geometry-grounded evidence package and passed to the downstream heuristic evidence generation and LLM reasoning steps.
\subsubsection{Heuristic Evidence and Scores}
The heuristic evidence generation step converts mesh evidence, topology evidence, orientation indicators, and lightweight task cues into rule-triggered fabrication evidence \((R,D,\Theta_0,M_{\mathrm{cand}}^{0})\). High support risk triggers support-related warnings, weak base contact triggers adhesion guidance, invalid topology triggers mesh-repair warnings, and high slenderness triggers stability warnings. Lightweight task cues further influence which risks and suggestions are emphasized. For example, clean-appearance tasks emphasize support scarring and underside quality, fit-related tasks emphasize dimensional accuracy and stability, and heat-related tasks emphasize material and shape-retention concerns. Based on these rule triggers, the layer also produces CAD-facing design hints \(D\), initial slicer-parameter baselines \(\Theta_0\), and keyword-based initial material candidates \(M_{\mathrm{cand}}^{0}\)~\cite{ahmad2022systematic,jiang2018support}.

\begin{figure*}[width=\textwidth,pos=t]
    \centering
    \includegraphics[width=0.95\linewidth]{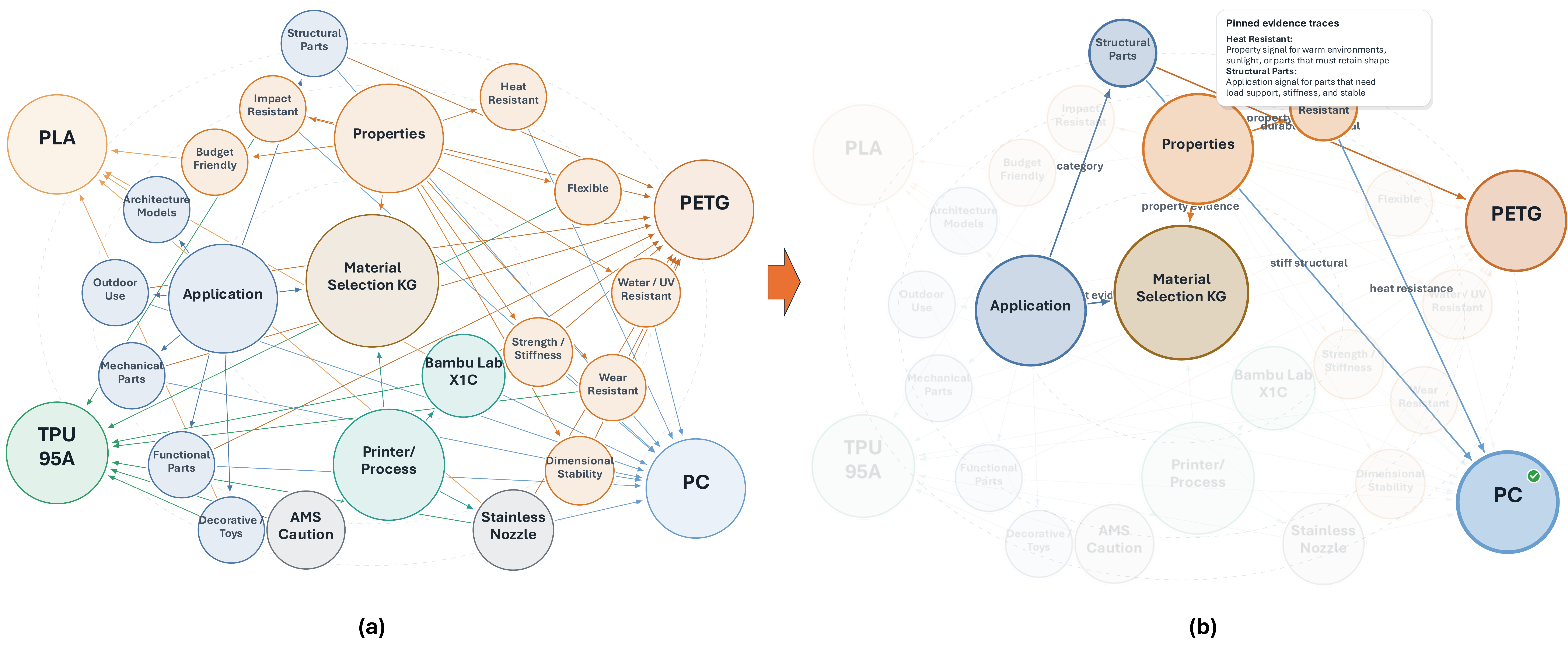}
    \caption{\textbf{Partial visualization and task-conditioned evidence tracing of the Task-Material-Process Knowledge Graph.} 
(a) The visualization shows the main node and edge types in the KG, including application nodes, material-property nodes, candidate-material nodes, and printer/process constraint nodes. These nodes organize task-related application contexts, material properties, material options, and printing-feasibility constraints. 
(b) Given a user task, the framework traces task-relevant application and property evidence through the KG to candidate materials. In this example, the task emphasizes structural use and heat resistant. PC receives the strongest converging evidence.}
    \label{fig:Knowledge_graph}
\end{figure*}
\subsection{Task-Material-Process Knowledge Graph}
\label{subsec:material_selection_kg}

The partial KG visualization in Figure~\ref{fig:Knowledge_graph} shows the main node and edge types, including application nodes, material-property nodes, printer/process constraint nodes, and candidate-material nodes. The complete Task-Material-Process KG contains 52 typed nodes, including 8 application nodes, 12 material-property nodes, 7 printer/process constraint nodes, and 25 candidate-material nodes. It also contains 347 typed edges, including 72 application-material edges, 96 property-material edges, and 179 printer/material feasibility edges. Application and material-property nodes provide the task-side semantic structure, representing application contexts and task-relevant material requirements. Candidate-material nodes encode available material options, while printer/process constraint nodes encode feasibility and process-related constraints under the printing setup. Together, these nodes provide traceable task-material-process knowledge for downstream LLM reasoning.

The Task-Material-Process KG was constructed through an expert-supervised, documentation-grounded, and LLM-assisted process. We first defined the graph schema, including the allowed node types, edge types, candidate-material scope, printer/process constraints, and scoring terms. Under this fixed schema, the LLM was used as an assistant to organize information \cite{bian2025llm} from official manufacturer material-printer documentation, group application and property categories, suggest missing task-property-material links, and initialize heuristic evidence strengths. All suggested nodes, links, and weights were normalized and checked by the experts before being used for retrieval and scoring. Therefore, the KG is not an unconstrained LLM-generated knowledge base; it is a schema-constrained material-printer knowledge layer used to provide traceable evidence for material recommendation. Table~\ref{tab:kg_responsibility} summarizes which parts of the KG were expert-defined and which parts were LLM-assisted.

\begin{table*}[width=\textwidth,pos=t]
\centering
\caption{Responsibility split in Task-Material-Process KG construction.}
\label{tab:kg_responsibility}
\begin{tabular}{p{0.20\textwidth}p{0.36\textwidth}p{0.36\textwidth}}
\hline
\textbf{KG component} & \textbf{Expert-defined} & \textbf{LLM-assisted} \\
\hline
Graph schema & Defined node types, edge types, and allowed relation types & No schema change \\
Candidate material scope & Fixed the material set used in the experiments & Did not introduce materials outside the allowed scope \\
Printer/process scope & Defined printer-feasibility fields and process-constraint categories & Helped organize compatibility information from documentation \\
Application nodes & Defined the application-node structure & Helped organize documentation-derived application categories and synonyms \\
Property nodes & Defined the material-property node structure & Helped extract and group material-property descriptions \\
Material nodes & Fixed candidate material identities and naming conventions & Helped normalize surface forms and synonyms \\
Edges & Defined allowed edge meanings, such as application-material, property-material, and printer-material feasibility & Suggested missing task-property-material links under the fixed schema \\
Edge weights & Defined scoring terms and normalization rules & Suggested initial heuristic evidence strengths \\
\hline
\end{tabular}
\end{table*}

\subsection{Geometry- and Knowledge-Grounded LLM Reasoning Layer}
\label{subsec:grounded_llm_reasoning}

The geometry- and knowledge-grounded LLM reasoning layer is the final task-driven reasoning layer of the framework. It receives the user task description \(T\), the geometry-grounded evidence package \(E_G\), and the KG-based material evidence package \(E_{KG}\), and generates the final structured recommendation report:
\begin{equation}
\label{eq:llm_output}
\begin{aligned}
O &= \mathcal{F}_{\mathrm{LLM}}(T,E_G,E_{KG}) \\
  &= \{y,M,R_T,\Theta,D_T,E\}.
\end{aligned}
\end{equation}
where \(y\) is the overall verdict, \(M\) is the final material recommendation, \(R_T\) is the task-relevant risk set, \(\Theta\) is the process-parameter guidance, \(D_T\) is the task-conditioned design guidance, and \(E\) is the rationale explaining the recommendation. This structure makes the output auditable because each component can be traced back to task intent, geometry-grounded evidence, or graph-retrieved material/process knowledge.

The geometry-grounded evidence \(E_G\) provides task-aware geometry risks, orientation guidance, parameter baselines, and design hints. The KG-based evidence \(E_{KG}\) provides ranked material candidates and supporting task-material-process evidence traces. By jointly using \(T\), \(E_G\), and \(E_{KG}\), the LLM performs constrained task-driven reasoning rather than unconstrained material or process recommendation.

\subsubsection{KG-Based Material Evidence Generation}

The KG-based material evidence is generated by a task-conditioned KG scoring function:
\begin{equation}
\label{eq:kg_evidence}
\begin{aligned}
E_{KG}
&=\mathcal{F}_{KG}(T,P,KG)=\{M_{\mathrm{rank}}, \Pi_{\mathrm{KG}}\}.
\end{aligned}
\end{equation}
where \(P\) is the fixed printer/process profile, \(M_{\mathrm{rank}}\) denotes the ranked material candidates and \(\Pi_{\mathrm{KG}}\) denotes the supporting KG evidence traces.
\begin{algorithm}[tbp]
\caption{Task-conditioned KG material scoring}
\label{alg:kg_retrieval}
\begin{algorithmic}[1]
\REQUIRE Task description \(T\), Task-Material-Process Knowledge Graph \(KG\), printer/process profile \(P\)
\ENSURE KG-based material evidence package \(E_{KG}\)
\STATE Activate task-relevant application and property nodes from \(T\).
\STATE Filter candidate materials using printer/process compatibility \(P\).
\FOR{each candidate material \(m\)}
    \STATE Retrieve KG evidence paths connecting activated application/property nodes to \(m\).
    \STATE Compute application and property evidence scores.
    \STATE Compute material difficulty and printer/process feasibility penalties.
    \STATE Compute the raw material score \(s_{\mathrm{raw}}(m)\).
    \STATE Store material \(m\), \(s_{\mathrm{raw}}(m)\), and supporting KG evidence traces.
\ENDFOR
\STATE Rank candidate materials by \(s_{\mathrm{raw}}(m)\).
\STATE Return \(E_{KG}\), including ranked material candidates and evidence traces.
\end{algorithmic}
\end{algorithm}

Algorithm~\ref{alg:kg_retrieval} summarizes the procedure used by \(\mathcal{F}_{KG}\) to produce \(E_{KG}\).
The material ranking in Algorithm~\ref{alg:kg_retrieval} uses an additive evidence score:
\begin{equation}
\label{eq:raw_score}
s_{\mathrm{raw}}(m)
=
s_{\mathrm{app}}(m)
+
s_{\mathrm{prop}}(m)
-
s_{\mathrm{diff}}(m)
-
s_{\mathrm{printer}}(m).
\end{equation}

The positive evidence terms are computed from task-activated application and property nodes:
\begin{equation}
\label{eq:positive_scores}
\begin{aligned}
s_{\mathrm{app}}(m)
&=
\sum_{a\in \mathcal{A}(T)}
\alpha_a w^{\mathrm{app}}_{a,m}, &s_{\mathrm{prop}}(m)=
\sum_{p\in \mathcal{P}(T)}
\alpha_p w^{\mathrm{prop}}_{p,m}.
\end{aligned}
\end{equation}

Here, \(\mathcal{A}(T)\) and \(\mathcal{P}(T)\) denote the application and property nodes activated by the task description. The coefficients \(\alpha\) denote activation strengths, and the weights \(w\) denote heuristic evidence strengths linking activated nodes to material \(m\). The penalty \(s_{\mathrm{diff}}(m)\) represents material printing difficulty, such as drying requirements, warping tendency, extrusion stability, and tuning sensitivity. The penalty \(s_{\mathrm{printer}}(m)\) represents printer/process infeasibility and acts as a feasibility constraint under the fixed experimental setup.

Materials with positive raw scores are ranked and passed to the LLM together with their supporting KG evidence traces. Therefore, the LLM does not perform unconstrained material selection. Instead, material recommendation is grounded by task-conditioned KG evidence and further integrated with the geometry-grounded evidence package \(E_G\) during final report generation.

\subsection{Post-Print Self-Improvement Module}
\label{subsec:feedback_loop}

The post-print self-improvement module provides an optional mechanism for refining future recommendations using post-print outcomes. After printing, the user can provide feedback \(F\), such as print success or failure, visible defects, dimensional-fit issues, support-removal difficulty, material behavior, or an overall satisfaction rating. The framework stores this feedback together with the task intent, geometry-grounded evidence, generated report, selected material, and process settings:
\begin{equation}
H=\{G,T,E_G,O,F,M,\Theta\},
\end{equation}
where \(M\) is the material used and \(\Theta\) denotes the process parameters used for printing.

The stored record \(H\) can be used to refine future recommendations at two levels. First, it provides case-level memory for later LLM reasoning, allowing the system to retrieve prior outcomes for similar tasks, materials, geometries, or failure modes. Second, the feedback record can support LLM-assisted refinement of both the geometry-grounded heuristics and the Task-Material-Process KG. On the geometry side, this module can adjust risk-trigger thresholds, orientation-scoring preferences, material-specific parameter baselines, and design-suggestion priorities. On the knowledge side, it can refine material-suitability scores, printing-difficulty penalties, task-property-material links, and retrieval preferences in the KG

\section{Experimental Design}

\subsection{Benchmark 3D Models}

Conventional printability studies often use all-in-one benchmark artifacts that combine many geometric features, such as overhangs, bridges, thin walls, holes, and fine details, into a single model~\cite{decker2015simplified,fudos2020characterization}. While such artifacts are useful for testing manufacturing limits, they are less suitable for our task-driven setting because multiple failure modes are mixed together and the intended task is usually implicit. Therefore, it becomes difficult to evaluate whether an assessment changes because of task intent or because of geometry alone.

\begin{figure}[pos=!t]
    \centering
    \includegraphics[width=\linewidth]{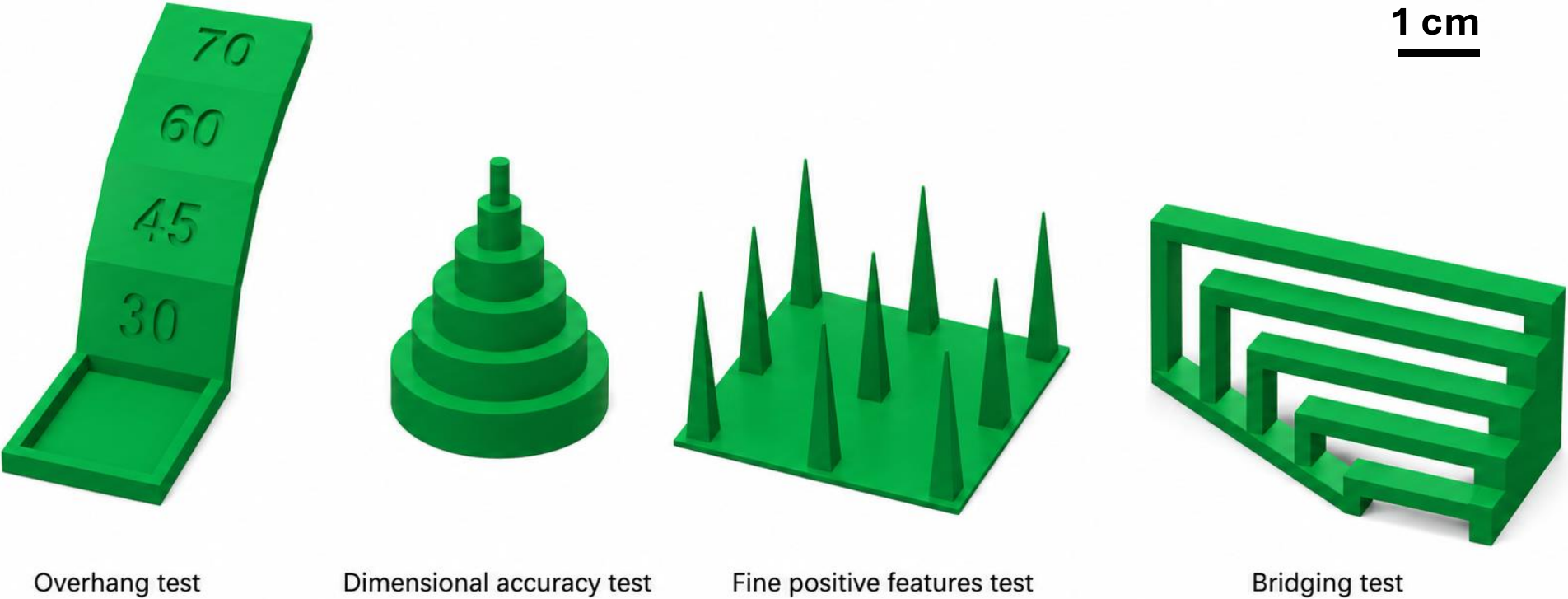}
    \caption{Benchmark models for task-driven printability evaluation.}
    \label{fig:benchmark_models}
\end{figure}

Instead, we use a set of focused STL benchmark models, each isolating a representative printability challenge. As shown in Figure~\ref{fig:benchmark_models} and Table~\ref{tab:benchmark_models}, the benchmark set includes an overhang test, a dimensional accuracy test, a fine positive feature test, and a bridging test. These models allow us to pair specific geometric risks with task-specific descriptions and evaluate whether the framework generates appropriate risks, explanations, and material/parameter/design suggestions.
\begin{table}[pos=htbp]
\centering
\caption{Benchmark 3D models.}
\label{tab:benchmark_models}
\renewcommand{\arraystretch}{1.18}
\setlength{\tabcolsep}{4pt}
\begin{tabular}{p{0.34\linewidth} p{0.56\linewidth}}
\toprule
\textbf{Benchmark model} & \textbf{Primary challenge} \\
\midrule
\arrayrulecolor{gray!20}

Overhang test &
Unsupported surfaces, support need, surface scarring, and orientation sensitivity. \\
\hline

Dimensional accuracy test &
Dimensional fidelity, tolerance, and fit-related accuracy. \\
\hline

Fine positive features test &
Small raised details, thin spikes, feature preservation, and local cooling sensitivity. \\
\hline

Bridging test &
Unsupported spans, sagging risk, cooling sensitivity, and bridge-related settings. \\

\arrayrulecolor{black}
\bottomrule
\end{tabular}
\end{table}

\subsection{Task Description Design}

Each focused STL benchmark is paired with two novice-style task descriptions, as shown in Table~\ref{tab:novice_task_descriptions}. The descriptions
are written in non-technical language and describe desired task
outcomes rather than explicit material properties or slicer settings.

For each description, we assign an implicit task priority and a reference
material as the ground-truth material label. The implicit priority
indicates the main requirement that the method should infer from
the user intent, while the reference material is used to evaluate
material recommendation accuracy.
\newcolumntype{P}[1]{>{\raggedright\arraybackslash}p{#1}}

\begin{table*}[width=\textwidth,pos=t]
\centering
\caption{Task descriptions, implicit task priorities, and reference material labels.}
\label{tab:novice_task_descriptions}
\renewcommand{\arraystretch}{1.22}
\setlength{\tabcolsep}{2.5pt}
\footnotesize
\begin{tabular}{@{}P{0.14\textwidth} P{0.54\textwidth}
                    P{0.20\textwidth} P{0.09\textwidth}@{}}
\toprule
\textbf{Benchmark Model} &
\textbf{Task Description} &
\textbf{Implicit Priority} &
\textbf{Reference Material} \\
\midrule
\arrayrulecolor{gray!20}

\multirow{2}{=}{Overhang test} &
I want this as a small indoor desk display used at room temperature.
It will not carry a load or be handled often. My main concern is a
clean underside without obvious sagging or support marks. &
Surface appearance; clean underside &
PLA \\
\cline{2-4}

&
I want to leave this part inside a parked car, where it may become hot.
It must remain rigid and keep the overhanging shape without bending or warping. &
Warm-environment use; shape retention &
PC \\
\hline

\multirow{2}{=}{Dimensional accuracy test} &
This part will be inserted into and removed from a matching rigid part many
times. The fit should remain firm and snug, and the edges should resist
chipping or cracking. I do not want the part to feel soft or rubbery. &
Fit reliability; firm shape &
PETG \\
\cline{2-4}

&
This part must stretch or compress when pushed onto another part and return
to its original shape after removal. It will be attached and removed
repeatedly and must not split or remain deformed. &
Slight flexibility; crack avoidance &
TPU 95A \\
\hline

\multirow{2}{=}{Fine positive features test} &
The small raised details will be touched and rubbed frequently and may
occasionally catch on other objects. They should remain clear and firm
without becoming soft, chipped, or cracked. &
Detail retention; firmness; chip/crack resistance &
PETG \\
\cline{2-4}

&
I want this to be part of a soft toy. The small raised parts should not
feel hard, sharp, or uncomfortable when touched. &
Softness; safety; hand feel &
TPU 95A \\
\hline

\multirow{2}{=}{Bridging test} &
I want this bridge-like part to look clean on the bottom. I do not want
the bottom surface to droop or look rough. &
Clean underside; low sagging &
PLA \\
\cline{2-4}

&
I want this bridge-like part to stay on the dashboard inside the car
and still keep its shape. It only needs to hold something light. &
Warm-environment use; shape retention &
PC \\

\arrayrulecolor{black}
\bottomrule
\end{tabular}
\end{table*}

\begin{table*}[width=\textwidth,pos=t]
\centering
\caption{Operational task-suitability tests and binary pass criteria.
Within each benchmark, the two rows follow the order of the corresponding
task descriptions in Table~\ref{tab:novice_task_descriptions}. The same
protocol was applied to all successfully fabricated samples from the
compared methods.}
\label{tab:task_suitability_protocols}
\renewcommand{\arraystretch}{1.22}
\setlength{\tabcolsep}{3pt}
\footnotesize
\begin{tabular}{@{}P{0.15\textwidth} P{0.47\textwidth}
                    P{0.34\textwidth}@{}}
\toprule
\textbf{Benchmark Model} &
\textbf{Test Procedure} &
\textbf{Pass Criterion} \\
\midrule
\arrayrulecolor{gray!20}

\multirow{2}{=}{Overhang test} &
After support removal, if supports are used, inspect the underside and
measure the maximum sag relative to the intended underside profile. &
No detached strands or major support-contact damage, and maximum
underside sag $\leq 0.5$\,mm. \\
\cline{2-3}

&
Expose the sample to 85$^\circ$C for 30\,min, cool it at room temperature
for 10\,min, and measure the permanent tip displacement. &
No cracking or visible warping, and permanent tip displacement
$\leq 2\%$ of the overhang length. \\
\hline

\multirow{2}{=}{Dimensional accuracy test} &
Perform 20 insertion and removal cycles using the same standardized mating
part, followed by a 500\,g retention test for 10\,s. &
The mating part remains retained, with no visible loosening, chipping,
or cracking. \\
\cline{2-3}

&
Perform 20 insertion and removal cycles using the same standardized mating part.
Measure the critical dimension before testing and 60\,s after the final cycle. &
No splitting or cracking, and dimensional recovery $\geq 95\%$ of the
initial value. \\
\hline

\multirow{2}{=}{Fine positive features test} &
Move a 1.0-mm-thick cardboard strip across the same raised-detail region
for 20 cycles under a 300\,g normal load. &
All target details remain distinguishable, with no visible chipping,
cracking, detachment, or permanent deformation. \\
\cline{2-3}

&
Compress the raised features to approximately 50\% of their initial height
for 10 cycles using a flat-ended probe. Measure feature height 60\,s after
the final cycle. &
Height recovery $\geq 90\%$, with no cracking, permanent collapse,
or sharp exposed edges. \\
\hline

\multirow{2}{=}{Bridging test} &
Inspect the bridge underside and measure the maximum mid-span sag relative
to the intended bridge profile. &
Mid-span sag $\leq 2\%$ of the bridge span, with no detached strands
or major surface discontinuities. \\
\cline{2-3}

&
Apply a 300\,g center load while exposing the sample to 85$^\circ$C
for 30\,min. Remove the load, cool for 10\,min, and measure the residual
mid-span deflection. &
No cracking, and residual mid-span deflection $\leq 2\%$ of the
bridge span. \\

\arrayrulecolor{black}
\bottomrule
\end{tabular}
\end{table*}

\subsection{Printer and Material Setup}
\label{Printer and Material Setup}
All experiments are conducted under a fixed printer profile using
a Bambu Lab X1 Carbon (X1C). Therefore, the framework does
not perform printer selection; instead, all material and parameter
recommendations are conditioned on the available X1C setup.

The material candidate set is limited to four laboratory-available materials: PLA, TPU 95A, PETG, and PC. As summarized in Table~\ref{tab:material_properties}, these materials cover distinct task-relevant properties, including ease of printing, flexibility, toughness, and heat resistance.

\begin{table}[pos=t]
\centering
\caption{Material properties}
\label{tab:material_properties}
\footnotesize
\setlength{\tabcolsep}{2.9pt}
\renewcommand{\arraystretch}{1.08}
\begin{tabular}{@{}lcccc@{}}
\toprule
\textbf{Material} & \makecell{\textbf{Printability}} & \textbf{Flexibility} & \textbf{Durability} & \makecell{\textbf{Heat resistance}} \\
\midrule
PLA     & High   & Low    & Medium & Low \\
PETG    & Medium & Medium    & Medium   & Medium \\
TPU 95A & Low & High   & High   & Medium \\
PC      & Low    & Low    & High   & High \\
\bottomrule
\end{tabular}
\end{table}

\subsection{Compared Methods / Baselines}

We compare four method settings. The rule-based baseline uses only the Geometry-Grounded Layer, without LLM reasoning or material-knowledge grounding. The pure LLM baseline uses only the LLM with the task description and the raw ASCII STL input, without the explicit geometry-grounded evidence package or material selection knowledge graph. The LLM + KG baseline adds the structured Task-Material-Process Knowledge Graph to the pure LLM setting, but still does not use the Geometry-Grounded Layer. Our method combines the Geometry-Grounded Layer and the LLM Layer with structured material/printer knowledge, enabling the LLM to reason over both computed geometry evidence and task-relevant material constraints. The self-improvement module is evaluated separately because it requires post-print feedback or expert correction.

For LLM-based settings, we evaluate three Gemini 2.5 backbones:
Flash-Lite, Flash, and Pro. The four methods
 are tested under the same backbone settings, so that the
comparison focuses on the effect of geometry and knowledge
grounding rather than the choice of LLM backbone.

\section{Evaluation Metrics Overview}
We evaluate the compared methods from four perspectives: printability evaluation, task suitability evaluation, material selection accuracy and expert evaluation of report quality. In addition, we assess the self-improvement module separately for the proposed framework. 

\subsection{Geometry-Level Printability Evaluation}
A trial is counted as successful if the generated recommendation is
executable and produces a physical print without a major
geometry-level failure. Recommendations that omit essential
information, such as material, orientation, or process parameters,
are counted as printability failures.

The observed Geometry-Level Printability rate is
\begin{equation}
\hat{p}_{\mathrm{print}}
=
\frac{N_{\mathrm{print}}}{N_{\mathrm{all}}},
\label{eq:printability_rate}
\end{equation}
where $N_{\mathrm{print}}$ is the number of successful prints and
$N_{\mathrm{all}}$ is the total number of validation trials.

The evaluation includes 96 trials from 8 task-conditioned scenarios,
4 methods, and 3 repetitions, giving $N_{\mathrm{all}}=24$ per
method. Of these trials, 75 proceeded to physical printing, while 21
non-executable recommendations were retained as printability failures.

\subsection{Task Suitability Evaluation}
\label{sec:task_suitability_eval}

Task suitability evaluation is conducted on successfully printed samples from the physical printability validation. A sample is entered into task-level testing only if it is successfully fabricated, because failed prints do not provide usable physical parts for downstream task suitability evaluation.

Each successfully printed sample is evaluated using the task-suitability test defined in Table~\ref{tab:task_suitability_protocols}. These tests assess underside quality and sag, heat-induced shape retention, repeated mating and retention, dimensional recovery after repeated insertion and removal, raised-feature abrasion resistance, feature recovery and touch safety, bridge underside sag, and residual bridge deflection under combined heat and load. A sample is counted as task-suitable only if it satisfies the corresponding binary pass criterion in Table~\ref{tab:task_suitability_protocols}.

We report two task-suitability metrics: the observed task-suitability
rate among successful prints, $\hat{p}_{(\mathrm{task}\mid\mathrm{print})}$,
and the observed end-to-end task-suitable rate,
$\hat{p}_{\mathrm{e2e}}$. They are computed as
\begin{align}
\hat{p}_{(\mathrm{task}\mid\mathrm{print})}
&=
\frac{N_{\mathrm{task}}}{N_{\mathrm{print}}},
\\
\hat{p}_{\mathrm{e2e}}
&=
\frac{N_{\mathrm{task}}}{N_{\mathrm{all}}},
\end{align}
where $N_{\mathrm{all}}$, $N_{\mathrm{print}}$, and
$N_{\mathrm{task}}$ denote the total trials, successful prints, and
task-suitable prints, respectively. In this study,
$N_{\mathrm{all}}=24$ for each method. If
$N_{\mathrm{print}}=0$,
$\hat{p}_{\mathrm{task}\mid\mathrm{print}}$ is reported as N/A.

\paragraph{Confidence intervals.}
We report marginal Wilson 95\% confidence intervals for
$\hat{p}_{\mathrm{print}}$ and $\hat{p}_{\mathrm{e2e}}$. Wilson
intervals are used because the outcomes are binary and include
boundary values near 0 or 1, where normal-approximation intervals may
be unreliable. For $\hat{p}=k/n$, the interval is
\begin{equation}
\mathrm{CI}_{\mathrm{Wilson}}
=
\frac{
\hat{p}+\frac{z^{2}}{2n}
\pm
z\sqrt{
\frac{\hat{p}(1-\hat{p})}{n}
+
\frac{z^{2}}{4n^{2}}
}
}{
1+\frac{z^{2}}{n}
},
\label{eq:wilson_interval}
\end{equation}
where $z=1.96$ and $n=24$. These intervals describe the precision of
the aggregate success-rate estimates under comparable evaluation
conditions, rather than instance-specific predictions.

\subsection{Material Selection Accuracy}
Since material choice is coupled with both geometry-level printability and task suitability, material-selection behavior provides important evidence for interpreting the physical validation results. We therefore evaluate material-selection accuracy and analyze the corresponding confusion patterns to identify where each method succeeds or fails in task-conditioned material reasoning.

Material selection accuracy measures whether the primary recommended material matches the reference material label in Table~\ref{tab:novice_task_descriptions}. To assess prediction stability, each of the eight task descriptions was evaluated five times, yielding 40 material predictions for each LLM-based method under each backbone. With three LLM-based methods and three Gemini 2.5 backbones, this resulted in 360 LLM-based predictions. The deterministic rule-based baseline was evaluated over the same 40 repeated cases for comparison, giving 400 material predictions in total.

\FloatBarrier
\subsection{Expert Evaluation of Report Quality}
Beyond geometry-level printability, task suitability, and material selection,the generated reports also explain why specific recommendations are made. Such explanations are important for novice users, but they are difficult to quantify with objective metrics. Therefore, we use expert evaluation to assess the quality of the generated reports.

Expert evaluation assesses each structured report based
on six output components(as Eq.~\eqref{eq:llm_output}): overall verdict, material recommendation,
task-relevant risks, parameter guidance, design guidance, and
explanation. This evaluation focuses on the quality and usefulness of the generated pre-print recommendations. It does not directly measure downstream functional performance, such as thermal deformation, snap-fit durability, surface softness, or long-term fit retention. The rubric of expert evaluation is shown in Table~\ref{tab:expert_rubric}.

\begin{table*}[width=\textwidth,pos=!t]
\centering
\caption{Expert evaluation rubric for structured reports.}
\label{tab:expert_rubric}
\renewcommand{\arraystretch}{1.20}
\setlength{\tabcolsep}{4pt}
\begin{tabular}{p{0.18\linewidth}p{0.35\linewidth}p{0.38\linewidth}}
\toprule
\textbf{Component} & \textbf{What it evaluates} & \textbf{Scoring criterion} \\
\midrule
\arrayrulecolor{gray!25}

Overall verdict $y$ &
Whether the report gives a correct and task-aware printability judgment. &
1: misleading or unsupported verdict; 5: accurate verdict that reflects geometry, task, and setup constraints. \\
\hline

Material $M$ &
Whether the recommended material is feasible and appropriate for the task. &
1: incompatible or task-irrelevant material; 5: correct, feasible, and well-matched material choice. \\
\hline

Risks $R_T$ &
Whether the report identifies task-relevant printability and functional risks. &
1: misses major risks or gives generic risks; 5: covers key risks and connects them to the task and geometry. \\
\hline

Parameters $\Theta$ &
Whether the process guidance is feasible and useful for the given printer/material setup. &
1: vague, infeasible, or unsafe parameters; 5: specific and practical parameters that address the identified risks. \\
\hline

Design guidance $D_T$ &
Whether the report provides useful CAD or geometry modification suggestions. &
1: no useful design guidance; 5: actionable design changes that directly improve printability or task suitability. \\
\hline

Explanation $E$ &
Whether the report clearly explains the reasoning behind the recommendation. &
1: unclear or disconnected explanation; 5: clear reasoning linking task intent, geometry evidence, material choice, parameters, and risks. \\

\arrayrulecolor{black}
\bottomrule
\end{tabular}
\end{table*}

Since expert scoring is time-consuming, we randomly select 10
cases from the 40 runs to reduce expert fatigue
and potential scoring errors. For each selected case, three experts
evaluate anonymized reports from the four compared methods. The reports were anonymized and randomly shuffled prior to evaluation so that the experts were blinded to the generating methods.

\FloatBarrier
\subsection{Post-Print Self-Improvement Module Evaluation}

The post-print self-improvement module is evaluated only for the proposed framework, because the baselines do not include a post-print refinement mechanism. The evaluation covers two types of problematic cases. We select 10 recommendation cases with material-selection errors or low expert scores to evaluate whether self-improvement improves material recommendation and report quality. We also apply the self-improvement module to the six failed physical print trials from the proposed framework to evaluate whether observed print failures can be corrected through regenerated recommendations.

For each selected case, post-print outcome information is added to the framework, and the recommendation is regenerated. The recorded information includes print outcomes, failure modes, material-related issues, and task-level suitability observations when available. The regenerated outputs are evaluated using the same criteria as the main evaluation, including material selection accuracy, printability, task suitability, and expert report quality.
\section{Results and Analysis}

\subsection{Geometry-Level Printability Evaluation}
Table~\ref{tab:physical_validation_results} shows that the proposed framework achieves the best overall printability performance among the compared methods. The most informative contrast is between LLM + KG and the proposed framework; although KG grounding improves material-related reasoning, it does not by itself produce complete and physically executable recommendations.

\begin{table}[pos=!t]
\centering
\caption{Printability and task-suitability evaluation over physical validation trials.}
\label{tab:physical_validation_results}
\renewcommand{\arraystretch}{1.15}
\setlength{\tabcolsep}{3.5pt}
\footnotesize
\begin{tabular}{@{}lccc@{}}
\toprule
\textbf{Method} &
$\boldsymbol{\hat{p}}_{\mathrm{print}}$ &
$\boldsymbol{\hat{p}}_{(\mathrm{task}\mid\mathrm{print})}$ &
$\boldsymbol{\hat{p}}_{\mathrm{e2e}}$ \\
\midrule

Rule-based
& 50.0\%
& 66.7\%
& 33.3\% \\

Pure LLM
& 37.5\%
& 55.6\%
& 20.8\% \\

LLM + KG
& 0.0\%
& N/A
& 0.0\% \\

Ours: LLM + KG + Geo
& \textbf{75.0\%}
& \textbf{88.9\%}
& \textbf{66.7\%} \\

\bottomrule
\end{tabular}
\end{table}

 LLM + KG often fails to generate essential guidance such as orientation and support-related decisions, which leads to poor printability despite improved material reasoning. This indicates that material knowledge alone is insufficient for printability assistance when geometry-grounded fabrication evidence is missing.

Pure LLM performs better than  LLM + KG in the printability evaluation, but this does not indicate stronger grounded reasoning. A more likely explanation is that pure LLM can still produce generic fabrication suggestions from its internal prior knowledge, which helps maintain a relatively high complete-recommendation rate even when those suggestions are not well matched to the specific geometry or intended use. Its advantage is therefore one of generic completeness rather than geometry-grounded correctness.

The proposed framework improves printability because it combines task-aware material reasoning with geometry-grounded evidence, including heuristic risk analysis and orientation guidance. This combination leads to more complete and more executable recommendations than those of the compared baselines. The remaining failures should be interpreted together with the material selection results, since some materials that are better aligned with the intended use are also more difficult to fabricate reliably. 

\begin{figure}[pos=!t]
\includegraphics[width=\linewidth]{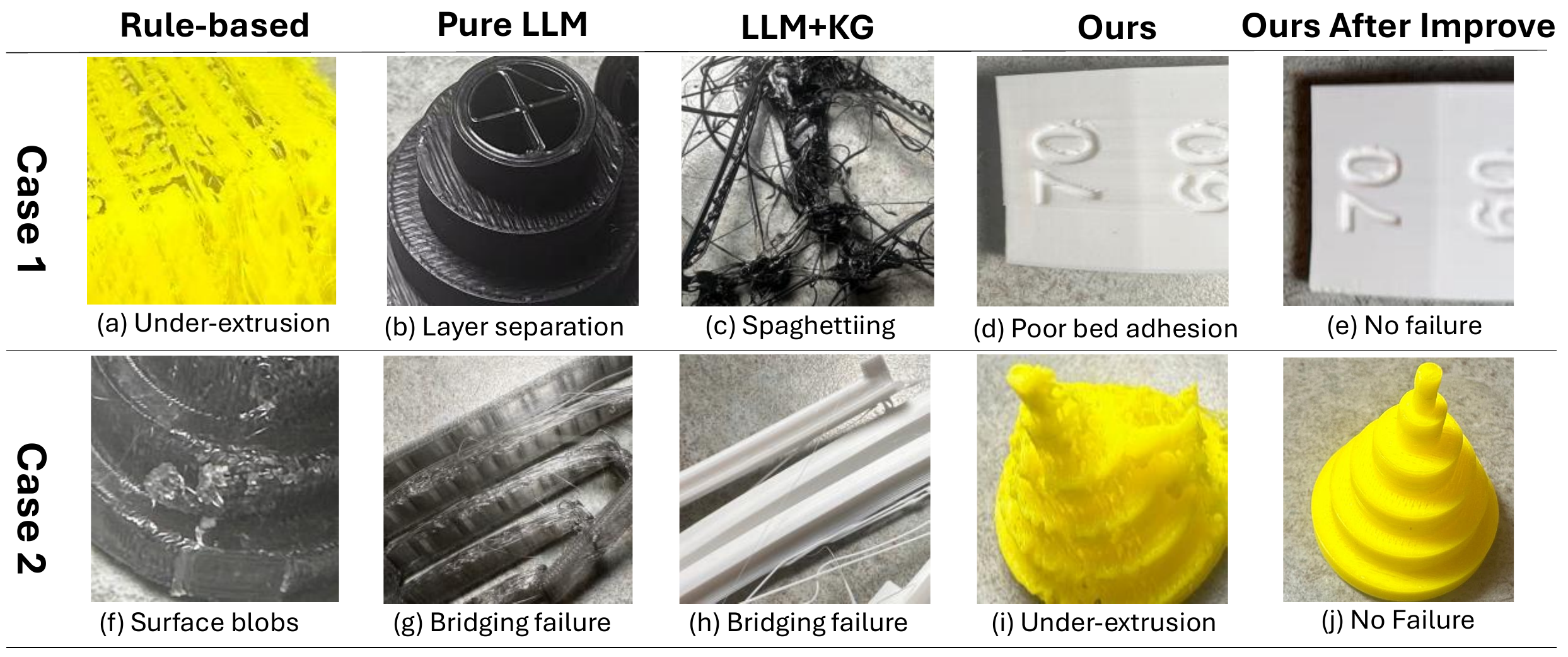}
\caption{Representative physical outcomes from geometry-level printability validation and post-print self-improvement. 
The first four columns show representative failure modes observed across the compared methods. 
The last column shows the corrected prints after applying the post-print self-improvement module to the failed cases.}
\label{fig:failure_cases}
\end{figure}

Figure~\ref{fig:failure_cases} shows representative geometry-level failure cases observed during physical validation, illustrating the types of fabrication failures counted in the printability evaluation.

\subsection{Task Suitability Evaluation}

Table~\ref{tab:physical_validation_results} also reports the
task-suitability results. Our method achieves the highest
task-suitability rate among successfully printed samples, with
$\hat{p}_{(\mathrm{task}\mid\mathrm{print})}=88.9\%~(16/18)$.
It also achieves the highest end-to-end task-suitable rate, with
$\hat{p}_{\mathrm{e2e}}=66.7\%~(16/24)$, compared with 33.3\%
for the rule-based baseline, 20.8\% for Pure LLM, and 0.0\%
for LLM + KG. These results indicate that the proposed framework
improves both physical printability and the likelihood that a
fabricated part satisfies its intended task requirement.

\begin{figure*}[width=\textwidth,pos=!t]
    \centering   \includegraphics[width=\textwidth]{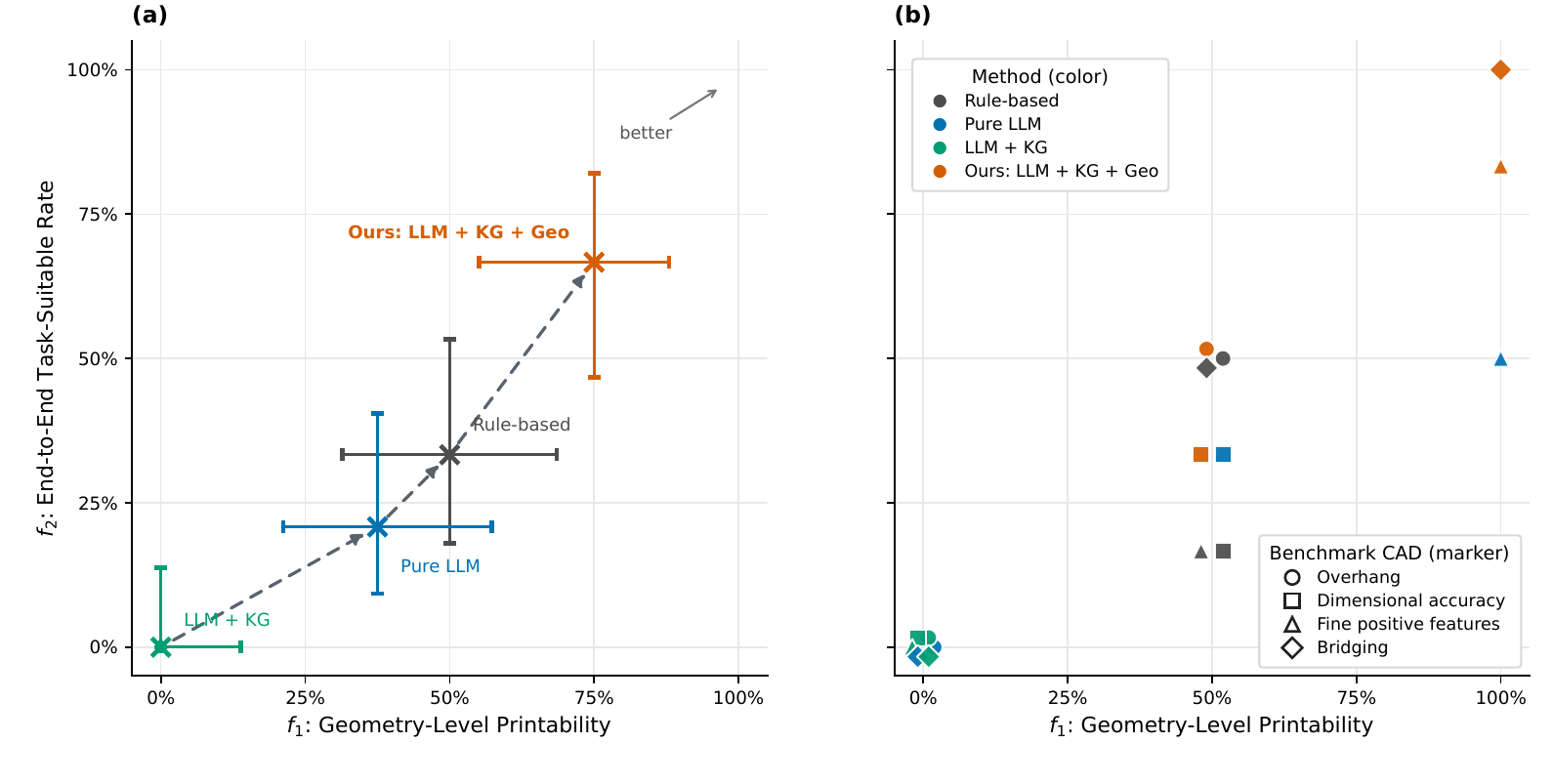}
    \caption{Joint comparison of geometry-level printability and task
suitability under Gemini 2.5 Flash-Lite. The horizontal axis reports
geometry-level printability, $f_1=\hat{p}_{\mathrm{print}}$, and the
vertical axis reports the end-to-end task-suitable rate,
$f_2=\hat{p}_{\mathrm{e2e}}$. In (a), each cross aggregates 24 trials
for one method, with horizontal and vertical error bars showing the
corresponding Wilson 95\% confidence intervals. These intervals
quantify uncertainty in the estimated success rates and indicate the
range of underlying performance compatible with the observations for
future comparable trials. In (b), each point aggregates six trials for one
method-benchmark CAD combination. Colors distinguish the four
methods, while marker shapes distinguish the four benchmark CAD
models. Exactly overlapping points are slightly offset for visibility.
Higher values on both axes indicate better joint performance.}
    \label{fig:joint_performance}
\end{figure*}

Figure~\ref{fig:joint_performance}(a) shows that our method occupies
the upper-right region and is the only non-dominated method based on
the observed rates. The dashed arrows indicate the direction of
stronger joint performance. The Wilson intervals describe the
uncertainty in the underlying rates for comparable trials, rather
than the outcome of an individual future trial.

Figure~\ref{fig:joint_performance}(b) reveals different
benchmark-level patterns. Rule-based remains at 50.0\% printability
across all four benchmark CAD models, while Pure LLM is more
benchmark-dependent, with printability distributed across 0.0\%,
50.0\%, and 100.0\%. All LLM~+~KG points remain at 0.0\%. In
contrast, our method maintains at least 50.0\% printability on every
benchmark CAD model, and three of the four models achieve an
end-to-end task-suitable rate of at least 50.0\%. The strongest
results are obtained for bridging, with 100\% on both objectives,
and fine positive features, with 100\% printability and 83.3\%
end-to-end task suitability.
\subsection{Material Selection Accuracy}
\begin{figure*}[width=\textwidth,pos=!t]
    \centering
    \includegraphics[width=0.70\linewidth]{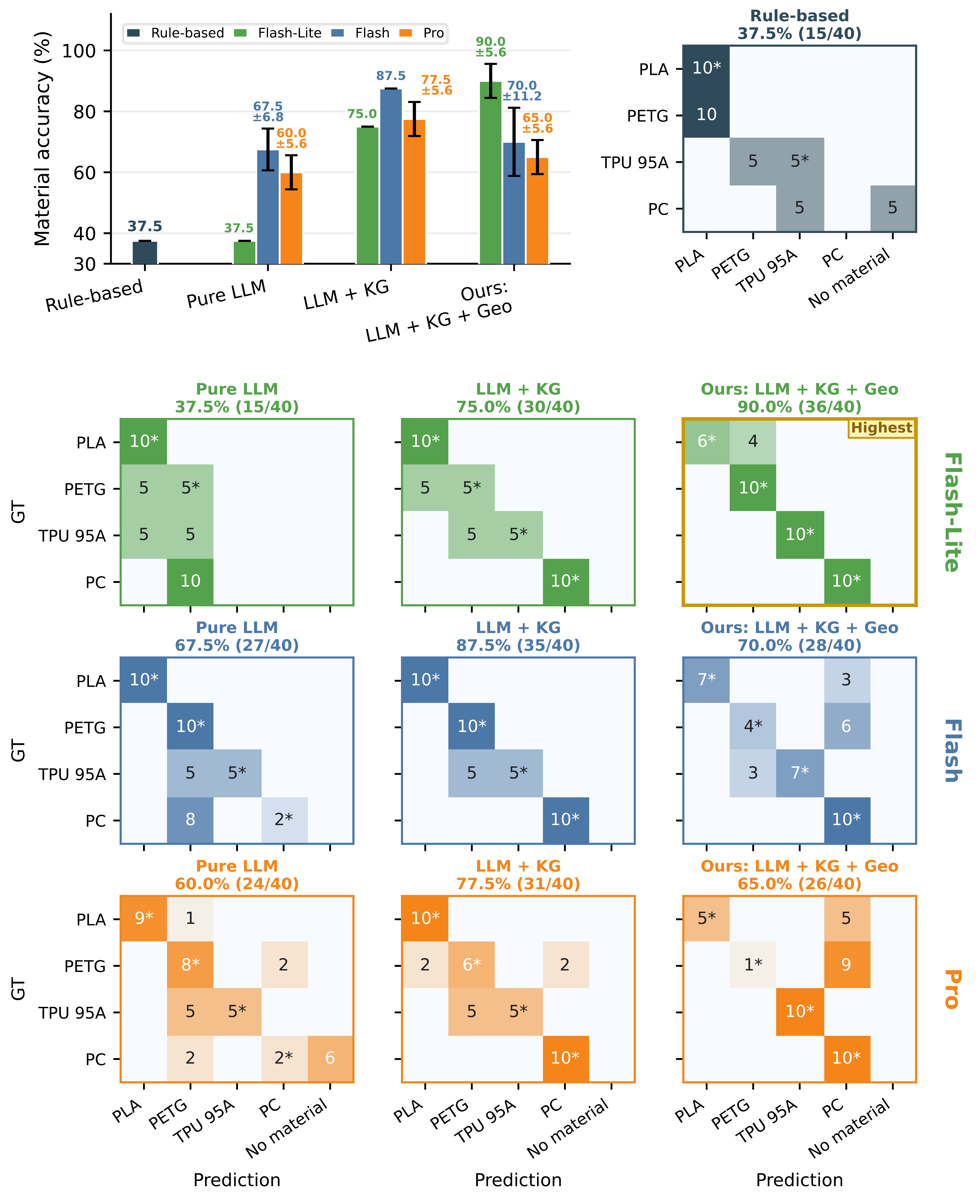}
    \caption{\textbf{Material selection accuracy and confusion matrices across evaluation methods.} The bar chart reports material recommendation accuracy under different methods across three Gemini model scales. Error bars indicate standard deviation across five runs. The confusion matrices show primary material predictions for each LLM-based method and model scale, where rows denote ground-truth materials and columns denote predicted primary materials. Asterisks mark correct predictions.}
    \label{fig:material acc}
\end{figure*}

Figure~\ref{fig:material acc} summarizes the material selection accuracy across methods and LLM backbones. The proposed framework achieves the best overall result, with the highest accuracy obtained by Gemini 2.5 Flash-Lite, where accuracy improves from 37.5\% under pure LLM prompting to 90.0\%. The confusion matrices also show more stable predictions, with fewer off-category outputs and fewer missing recommendations.

The progression across methods is informative. Under Gemini 2.5 Flash-Lite, pure LLM prompting performs similarly to the rule-based baseline, suggesting that free-form LLM reasoning alone does not provide sufficient grounded information for reliable material prediction. Adding the KG substantially improves accuracy, indicating that structured material knowledge is a major source of evidence for this task. The proposed framework further improves upon pure LLM + KG, suggesting that geometry-grounded rule-based evidence and heuristic priors provide additional complementary information. Together, these results support the framework design, in which KG grounding strengthens material reasoning and geometry-grounded evidence further refines task-conditioned material selection.

The remaining errors of the proposed framework are mainly concentrated in PLA and PETG confusion, rather than larger mismatches involving TPU 95A or PC. This pattern is understandable because PLA and PETG are relatively similar within the evaluated material set and can both satisfy some rigid-part requirements, although PETG is typically tougher while PLA is easier to print. This suggests residual ambiguity between nearby material choices rather than arbitrary failure, while materially distinct categories remain more reliably separated.

A further trend can also be observed across the evaluated LLM backbones. In the pure LLM and pure LLM + KG settings, Gemini Flash generally performs better than Flash-Lite, which may indicate that the larger model contains richer internal knowledge relevant to 3D printing and material selection. However, this trend does not continue from Flash to Pro. Instead, performance decreases at the largest scale, which may reflect an inverse-scaling~\cite{mckenzie2023inverse, wei2023inverse, michaelov2023rarely} effect in the present evaluation setting. A possible explanation is that larger models are more likely to generate alternative but still plausible material choices based on broader prior knowledge, even when the evaluation uses a fixed reference material label \cite{yang2025llm}.

\subsection{Expert Evaluation of Report Quality}

\begin{figure*}[width=\textwidth,pos=!t]
    \centering
    \includegraphics[width=\linewidth]{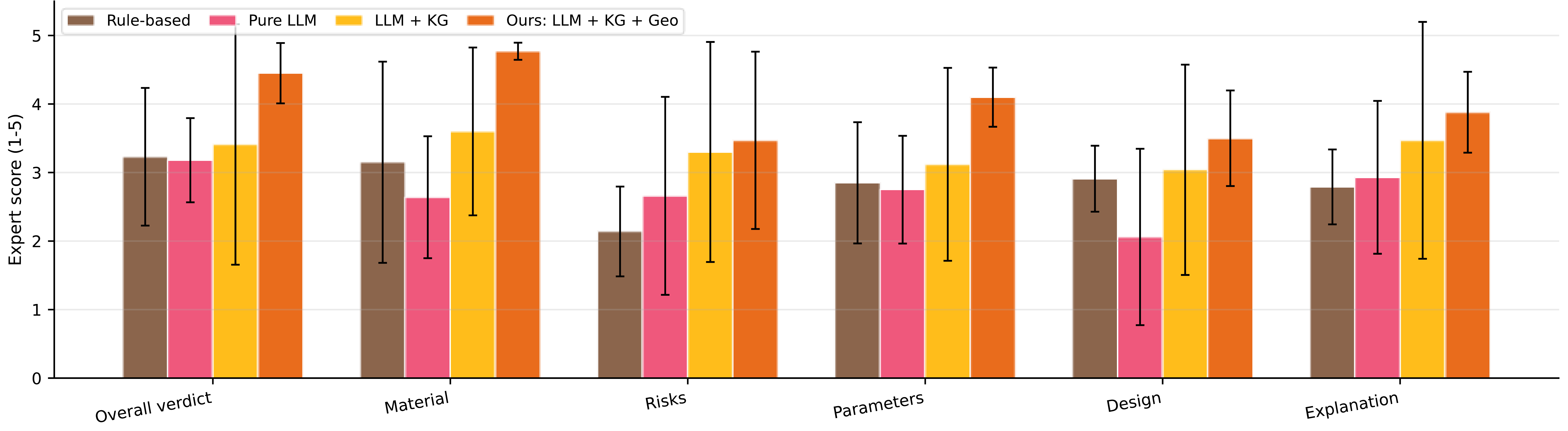}
    \caption{Expert evaluation of structured printability reports.}
    \label{fig:expert_score}
\end{figure*}

Figure~\ref{fig:expert_score} shows the expert
evaluation results across the six structured report components. The
proposed framework achieves the highest mean score in all
components, indicating that the generated reports are more complete,
task-aware, and actionable than those produced by the three baselines.

The rule-based baseline receives lower scores because it mainly
relies on local geometry heuristics. It can provide basic printability
warnings, but it has limited ability to infer the user's functional
intent, select task-appropriate materials, or explain trade-offs among
material, geometry, and process settings. The pure LLM baseline
improves over the rule-based baseline in most components, especially
in verdict generation and explanation. However, its performance is
less stable in material selection and design guidance, suggesting that
free-form LLM reasoning may produce plausible but insufficiently
grounded recommendations.

The proposed framework performs best in material recommendation,
task-relevant risk identification, parameter guidance, and explanation
quality. These improvements show the benefit of combining
geometry-grounded evidence with structured material and printer
knowledge. In particular, the higher scores for risks and parameters
suggest that the framework can better connect task intent with
geometry-dependent failure modes and feasible slicer-level actions.

\begin{figure*}[width=\textwidth,pos=!t]
    \centering
    \includegraphics[width=\linewidth]{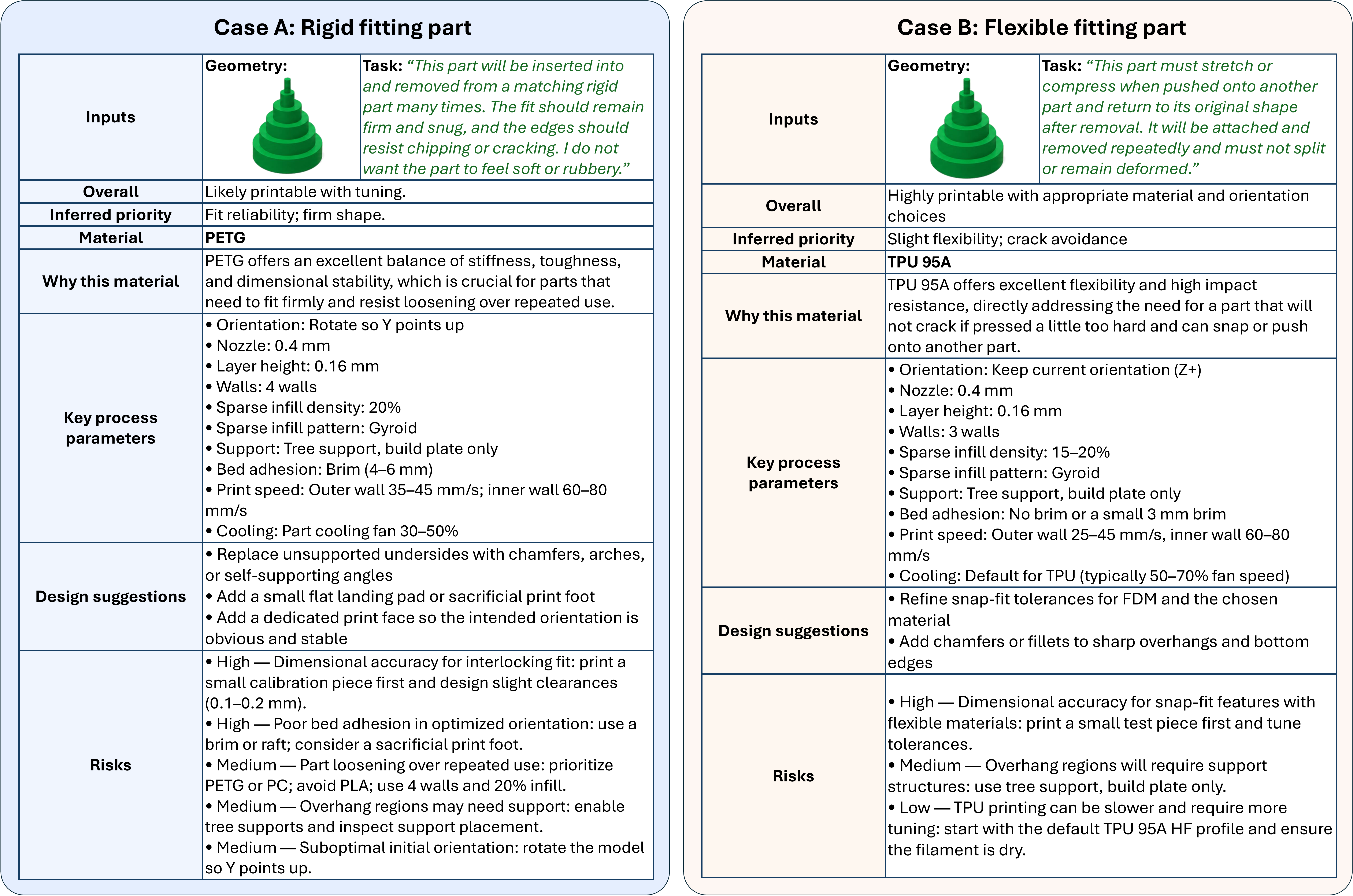}
    \caption{\textbf{Same geometry under different task case study.} The same STL geometry is evaluated under two novice-style task descriptions with different functional requirements. For the rigid fitting task, the framework infers fit reliability and firm shape as the main priorities and recommends PETG with higher wall count, moderate infill, brim-assisted adhesion, and a Y-up candidate orientation to support stiffness and repeated-use stability. For the flexible fitting task, the framework instead prioritizes slight flexibility and crack avoidance, recommending TPU 95A with the current Z+ orientation, lower infill, and TPU-specific tuning. This comparison shows that the proposed framework adapts material, parameter, design, and risk recommendations according to task intent rather than geometry alone.}
    \label{fig:same_geo}
\end{figure*}

In addition to the aggregate expert scores, Figure~\ref{fig:same_geo}
shows a representative same-geometry different-task case. The same
dimensional-accuracy STL leads to different recommendations under rigid
and flexible fitting intents: PETG and higher-stiffness settings for the
rigid fitting case, but TPU 95A and more flexible tuning for the snap-fit
case. This example qualitatively illustrates the task-conditioned behavior
measured by the  evaluation.

\subsection{Post-Print Self-Improvement Module Evaluation}

\begin{figure}[width=\columnwidth,pos=h]
    \centering
    \includegraphics[width=1\columnwidth,pagebox=cropbox]{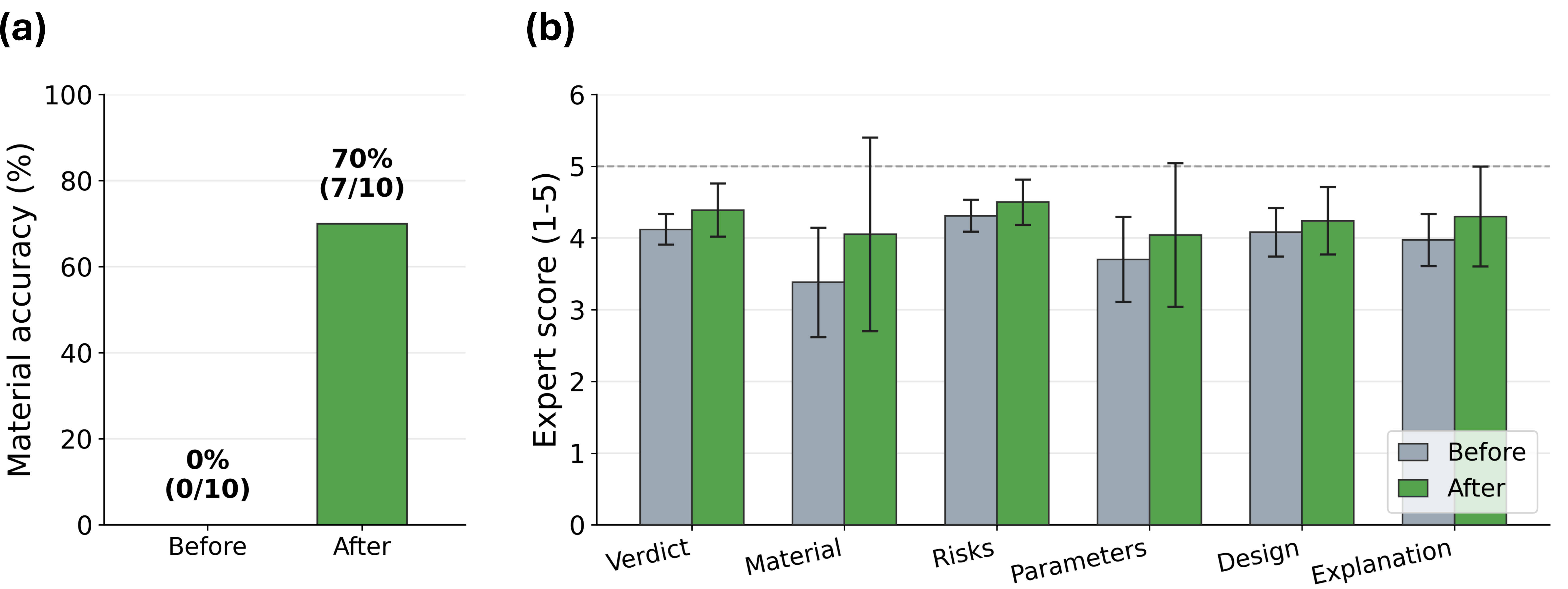}
    \caption{Effect of self-improvement module.}
    \label{fig:After_feedback}
\end{figure}

Figure~\ref{fig:After_feedback}(a) shows that material recommendation accuracy on the 10 selected problematic recommendation cases improves from 0/10 before self-improvement to 7/10 after self-improvement. Figure~\ref{fig:After_feedback}(b) shows that expert scores improve across all six structured report components, indicating that self-improvement improves the coherence and usefulness of the regenerated reports.

For physical validation, the self-improvement module is applied to the six failed print trials from the proposed framework. These failures correspond to two task-conditioned benchmark scenarios with three repeated trials each. After self-improvement, the failed trials are corrected, increasing the proposed framework's printability from 75.0\% to 100.0\%. Since the corrected printed samples also pass the corresponding task-suitability tests, the end-to-end task-suitable rate $\hat{p}_{\mathrm{e2e}}$ also improves from 66.7\% to 91.7\%.

Figure~\ref{fig:failure_cases} shows two representative physical outcomes before and after self-improvement. These examples qualitatively illustrate how observed post-print failures can be used to improve subsequent recommendations.

\section{Discussion}
The results show that task-driven printability assistance requires both geometry-grounded fabrication evidence and task-material-process knowledge. Geometry evidence makes the recommendation physically executable, while structured material/printer knowledge helps align the recommendation with the intended use. The weak printability of the LLM + KG baseline further indicates that material knowledge alone is insufficient without orientation, support, adhesion, and other geometry-dependent evidence.

The remaining failures reflect the trade-off between task suitability and fabrication difficulty. Materials that better satisfy functional requirements, such as flexibility or heat resistance, can also introduce higher printing risks. The current evaluation is limited to four focused STL models and four laboratory-available materials, so broader validation is needed across more geometries, printers, materials, and downstream functional tests.

For a single Gemini 2.5 Flash-Lite request, the geometry-grounded layer takes 0.003 s, KG-based material reasoning takes 5.350 s, and LLM inference takes 40.518 s, resulting in an end-to-end runtime of 45.871 s. The main bottleneck is LLM inference, while local geometry analysis adds negligible overhead. 
\section{Conclusion}

This paper presented a task-driven 3D printability assistance framework that combines geometry-grounded analysis, structured material/printer knowledge, and constrained LLM reasoning. Experiments show improved material selection, physical printability, task suitability, and expert-rated report quality compared with rule-based, pure LLM, and LLM + KG baselines. The post-print self-improvement evaluation further suggests that observed failures can refine subsequent recommendations, demonstrating the feasibility of geometry- and knowledge-grounded LLM reasoning for interactive, task-aware printability assistance.

\section*{Declaration of Generative AI Use}

ChatGPT (OpenAI) was used solely to improve the language and readability of this manuscript. The authors reviewed all revisions and take full responsibility for the content.

\section*{Online Supplementary Material}
Representative interactive outputs and case demonstrations are
available on the
\href{https://anonymous.4open.science/api/repo/CAD-paper-ano-D41D/file/index.html}
{project page}.


\balance
\bibliographystyle{cas-model2-names}
\bibliography{refs}

\end{document}